\documentclass[conference]{IEEEtran}
\IEEEoverridecommandlockouts
\usepackage{cite}

\ifCLASSINFOpdf
   \usepackage[pdftex]{graphicx}
\else
   \usepackage[dvips]{graphicx}
\fi
\usepackage[cmex10]{amsmath}
\usepackage{algorithm}
\usepackage{algorithmic}

\usepackage{array}

\usepackage{mdwmath}
\usepackage{mdwtab}
\usepackage[caption=false,font=footnotesize]{subfig}
\usepackage{bm}

\usepackage{fixltx2e}
\usepackage{url}

\usepackage{tabularx}
\usepackage{multirow}
\usepackage{booktabs}
\usepackage{amsfonts}
\usepackage{threeparttable}

\begin{document}
%

\title{Evolutionary Multitasking for Multiobjective Continuous Optimization: Benchmark Problems, Performance Metrics and Baseline Results}

\author{\IEEEauthorblockN{Yuan Yuan\IEEEauthorrefmark{1},
Yew-Soon Ong\IEEEauthorrefmark{1}, Liang Feng\IEEEauthorrefmark{2}, A.K. Qin\IEEEauthorrefmark{3}, Abhishek Gupta\IEEEauthorrefmark{1}, Bingshui Da\IEEEauthorrefmark{1},\\
Qingfu Zhang\IEEEauthorrefmark{4}, Kay Chen Tan\IEEEauthorrefmark{5}, Yaochu Jin\IEEEauthorrefmark{6}, and Hisao Ishibuchi\IEEEauthorrefmark{7}}
\IEEEauthorblockA{\IEEEauthorrefmark{1} School of Computer Science and Engineering,
Nanyang Technological University, Singapore\\
\IEEEauthorrefmark{2} College of Computer Science, Chongqing University, China \\
\IEEEauthorrefmark{3} School of Science, RMIT University, Australia  \\
\IEEEauthorrefmark{4} Department of Computer Science, City University of Hong Kong, Hong Kong  \\
\IEEEauthorrefmark{5} Department of Electrical and Computer Engineering, National University of Singapore, Singapore \\
\IEEEauthorrefmark{6} Department of Computer Science, University of Surrey, United Kingdom \\
\IEEEauthorrefmark{7} Department of Computer Science and Intelligent Systems, Osaka Prefecture University, Japan \\
\textbf{Technical Report}}}

\maketitle

\section{Introduction}
\label{sec-Introduction}

With the explosion in the variety and volume of incoming information streams,
it is very desirable that the intelligent systems or algorithms
are capable of efficient multitasking. The evolutionary algorithms (EAs)
are a kind of population-based search algorithms, which have the inherent ability
to handle multiple optimization tasks at once.
By exploiting this characteristic of EAs,
evolutionary multitasking \cite{gupta2016multifactorial,ong2016evolutionary} has become
a new paradigm in evolutionary computation (EC),
which signifies a multitasking search involving multiple optimization tasks at a time, with each task contributing a unique factor influencing the evolution of a single population of individuals.
The first tutorial
on evolutionary multitasking (``Evolutionary Multitasking and Implications for Cloud Computing'') was presented in IEEE Congress on Evolutionary Computation (CEC) 2015, Sendai, Japan.
Since then, a series of studies on this topic have been conducted
from various perspectives, which not only show the potentials of evolutionary
multitasking \cite{gupta2015evolutionary,gupta2016multiobjective,da2016evolutionary,chandra2016evolutionary,sagarna2016concurrently,yuan2016evolutionary} but also explain why this paradigm works \cite{gupta2016measuring,gupta2016genetic} to some extent.

Evolutionary multiobjective optimization (EMO) \cite{deb2001multi,coello2002evolutionary,tan2006multiobjective} has
always been a popular topic in the field of EC over the past 20 years. The reason mainly lies in
the following two aspects. On the one hand, many optimization problems in real-world applications
can be formulated as multiobjective optimization problems (MOPs)
in essence. On the other hand, EAs are able to approximate the whole Pareto set (PS) or Pareto front (PF) of a MOP in
a single run due to the population-based nature, which are regarded as quite suitable for solving MOPs.
Until now, plenty of EMO techniques have been developed, and they can be roughly
classified into three categories: Pareto dominance-based algorithms \cite{corne2001pesa,zitzler2001spea2,deb2002fast}, decomposition-based algorithms \cite{zhang2007moea,deb2014evolutionary,yuan2016new,yuan2016balancing},
and indicator-based algorithms \cite{zitzler2004indicator,beume2007sms,bader2011hype}.

Since evolutionary multitasking has shown promise in solving multiple tasks simultaneously and
EMO is one of the focused topics in EC, there may be noteworthy implications of understanding the behavior of algorithms at the intersection of the two paradigms.
The combination of the two different paradigms is referred to herein as \emph{multiobjective multifactorial optimization} (MO-MFO),
which means that a number of multiobjective optimization tasks are solved
simultaneously by evolving a single population of individuals using EAs.
MO-MFO has been first discussed in \cite{gupta2016multiobjective}, where proof of concept results from two synthetic benchmarks and a real-world case study were presented.

In this report, we suggest nine test problems for MO-MFO,
each of which consists of two multiobjective optimization tasks that need to be solved simultaneously.
The relationship between tasks varies between different test problems,
which would be helpful to have a comprehensive evaluation of the MO-MFO algorithms.
It is expected that the proposed test problems will germinate progress the field of the MO-MFO research.

The rest of this report is outlined as follows. Section \ref{sec-Definitions of Test Problems}
presents the definitions of the proposed test problems.
Section \ref{sec-Performance Evaluation} describes how to
evaluate the performance of the algorithm on the test problems.
Section \ref{sec-Baseline Results} provides the baseline results obtained by a MO-MFO algorithm (i.e., MO-MFEA \cite{gupta2016multiobjective}) and its
underlying basic MOEA (i.e., NSGA-II \cite{deb2002fast}) on the proposed benchmark problems.

\section{Definitions of Benchmark Problems}
\label{sec-Definitions of Test Problems}

In the previous studies on evolutionary (single-objective) multitasking \cite{gupta2016multifactorial,ong2016evolutionary,gupta2016measuring},
it is found that the degree of the intersection of the global optima and the similarity in the fitness landscape
are two important ingredients that lead to the complementarity between different optimization tasks.
Motivated by this, the characteristics of multiobjective optimization tasks in the
test problems to be presented are focused on the $q(\mathbf{x})$ function, and the relationship between two multiobjective
optimization tasks in a test problem can be translated into
the relationship between their respective $q(\mathbf{x})$ functions, which is based on
the fact that the Pareto optimal solutions of a multiobjective task are achieved iff.
its corresponding $q(\mathbf{x})$ reaches the global minimum.

Suppose that $\mathbf{x}_{1}^{*}$ and $\mathbf{x}_{2}^{*}$ are the global minima
of $q(\mathbf{x})$ functions in the two multiobjective tasks $T_{1}$ and $T_{2}$, respectively.
Each dimension of $\mathbf{x}_{1}^{*}$ and $\mathbf{x}_{2}^{*}$ is
then normalized to the same range $[0,1]$, obtaining $\mathbf{\bar{x}}_{1}^{*}$ and $\mathbf{\bar{x}}_{2}^{*}$.
If $\mathbf{\bar{x}}_{1}^{*}=\mathbf{\bar{x}}_{2}^{*}$, we say that the global minima
are \emph{complete intersection}; if there exists no dimension where the values of $\mathbf{\bar{x}}_{1}^{*}$ and $\mathbf{\bar{x}}_{2}^{*}$
are equal, we call that \emph{no intersection}; all the other relationships between $\mathbf{\bar{x}}_{1}^{*}$ and $\mathbf{\bar{x}}_{2}^{*}$
are referred to as \emph{partial intersection}.

For computing the similarity between the fitness landscape of $q(\mathbf{x})$ functions, 1,000,000 points are randomly sampled in the unified search space \cite{gupta2016multifactorial},
then the Spearman's rank correlation coefficient between $q(\mathbf{x})$ is calculated as the similarity.
The similarity lying in $(0, 1/3], (1/3,2/3],(2/3,1]$ is regarded as high, medium, and low, respectively.

We consider three degrees of the intersection of the global minima, i.e., complete, partial and no intersection, and within each category,
three categories of similarity in the fitness landscape, i.e., high, medium, and low similarity. Accordingly, there are
nine test problems in total.
Note that many practical settings give rise to a third condition for categorizing potential multitask optimization settings, namely, based on the phenotypic overlap of the decision variables \cite{ong2016evolutionary}. To elaborate, a pair of variables from distinct tasks may bear the same semantic (or contextual) meaning, which leads to the scope of knowledge transfer between them. However, due to the lack of substantial contextual meaning in the case of synthetic benchmark functions, such a condition for describing the similarity/overlap between tasks is not applied in this technical report.

The definitions of the proposed test problems are described in detail as follows.
 Note that there are five shift vectors ($\mathbf{s}_{cm2}, \mathbf{s}_{ph2}, \mathbf{s}_{pm1}, \mathbf{s}_{pl2}, \mathbf{s}_{nl1}$) and four rotation matrixes
($\mathbf{M}_{cm2}, \mathbf{M}_{pm1}, \mathbf{M}_{pm2}, \mathbf{M}_{nm2}$) involved, whose data can be available online\footnotemark[1]\footnotetext[1]{The data of rotation matrixes and shift vectors in the test problems can be downloaded from \\ https://drive.google.com/open?id=0B8WAZ9HjQsUSdUY5UzBLN0NPd2M.}:

1) \textbf{Complete Intersection with High Similarity (CIHS)}

The first multiobjective task $T_{1}$ is defined as follows:
\begin{equation}\label{eq-t11}
\begin{split}
& \min f_{1}(\mathbf{x}) = q(\mathbf{x})\cos(\frac{\pi x_{1}}{2}), \\
& \min f_{2}(\mathbf{x}) = q(\mathbf{x})\sin(\frac{\pi x_{1}}{2}), \\
& q(\mathbf{x}) = 1 + \sum_{i=2}^{n}x_{i}^{2}, \\
& x_{1} \in [0,1], x_{i} \in [-100,100], i = 2,3,\ldots, n \\
\end{split}
\end{equation}

The second multiobjective task $T_{2}$ is defined as follows:
\begin{equation}\label{eq-t12}
\begin{split}
& \min f_{1}(\mathbf{x})= x_{1},\\
& \min f_{2}(\mathbf{x})= q(\mathbf{x})(1-(\frac{x_{1}}{q(\mathbf{x})})^{2}), \\
&   q(\mathbf{x}) =1+ \frac{9}{n-1} \sum_{i=2}^{n}|x_{i}|  \\
& x_{1} \in [0,1], x_{i} \in [-100,100], i = 2,3,\ldots, n \\
\end{split}
\end{equation}

For this test problem, the number of decision variables $n$ is set to 50
for both $T_{1}$ and $T_{2}$, and the similarity between $T_{1}$
and $T_{2}$ is 0.97.
The PS and PF of
$T_{1}$ are given as follows:
\begin{equation}\label{eq-pf11}
\begin{split}
& PS:  x_{1} \in [0,1],  x_{i}=0, i= 2,3,\ldots, 50 \\
& PF: f_{1}^{2}+ f_{2}^{2}=1, f_{1} \geq 0, f_{2} \geq 0\\
\end{split}
\end{equation}
The PS and PF of
$T_{2}$ are given as follows:
\begin{equation}\label{eq-pf12}
\begin{split}
& PS:  x_{1} \in [0,1],  x_{i}=0, i= 2,3,\ldots, 50 \\
& PF: f_{2}=1-f_{1}^{2}, 0 \leq f_{1} \leq 1\\
\end{split}
\end{equation}
\\
2) \textbf{Complete Intersection with Medium Similarity (CIMS)}

The first multiobjective task $T_{1}$ is defined as follows:
\begin{equation}\label{eq-t21}
\begin{split}
& \min f_{1}(\mathbf{x})= x_{1},\\
& \min f_{2}(\mathbf{x})= q(\mathbf{x})(1-(\frac{x_{1}}{q(\mathbf{x})})^{2}), \\
&  q(\mathbf{x})= 1+\sum_{i=2}^{n-1} (100(x_{i}^{2}-x_{i+1})^{2} + (1-x_{i})^{2}),\\
& x_{1} \in [0,1], x_{i} \in [-5,5], i = 2,3,\ldots, n \\
\end{split}
\end{equation}

The second multiobjective task $T_{2}$ is defined as follows:
 \begin{equation}\label{eq-t22}
\begin{split}
& \min f_{1}(\mathbf{x}) = q(\mathbf{x})\cos(\frac{\pi x_{1}}{2}), \\
& \min f_{2}(\mathbf{x}) = q(\mathbf{x})\sin(\frac{\pi x_{1}}{2}), \\
& q(\mathbf{x}) =1+ \frac{9}{n-1} \sum_{i=2}^{n}|z_{i}|,  \\
& (z_{2}, z_{3}, \ldots z_{n})^{\mathrm{T}} = \mathbf{M}_{cm2}((x_{2}, x_{3}, \ldots, x_{n})^{\mathrm{T}} - \textbf{s}_{cm2}), \\
& x_{1} \in [0,1], x_{i} \in [-5,5], i = 2,3,\ldots, n \\
\end{split}
\end{equation}

For this test problem, the number of decision variables $n$ is set to 10
for both $T_{1}$ and $T_{2}$, and the similarity between $T_{1}$
and $T_{2}$ is 0.52.
The PS and PF of
$T_{1}$ are given as follows:
\begin{equation}\label{eq-pf21}
\begin{split}
& PS:  x_{1} \in [0,1],  x_{i}=1, i= 2,3,\ldots, 10 \\
& PF: f_{2}=1-f_{1}^{2}, 0 \leq f_{1} \leq 1\\
\end{split}
\end{equation}
The PS and PF of
$T_{2}$ are given as follows:
\begin{equation}\label{eq-pf22}
\begin{split}
& PS: x_{1} \in [0,1], (x_{2}, x_{3}, \ldots, x_{10})^{\mathrm{T}} = \textbf{s}_{cm2} \\
& PF: f_{1}^{2}+ f_{2}^{2}=1, f_{1} \geq 0, f_{2} \geq 0\\
\end{split}
\end{equation}
\\


3) \textbf{Complete Intersection with Low Similarity (CILS)}

The first multiobjective task $T_{1}$ is defined as follows:
\begin{equation}\label{eq-t31}
\begin{split}
& \min f_{1}(\mathbf{x}) = q(\mathbf{x})\cos(\frac{\pi x_{1}}{2}), \\
& \min f_{2}(\mathbf{x}) = q(\mathbf{x})\sin(\frac{\pi x_{1}}{2}), \\
& q(\mathbf{x}) = 1+\sum_{i=2}^{n} (x_{i}^{2} - 10 \cos(2 \pi x_{i}) + 10)  \\
& x_{1} \in [0,1], x_{i} \in [-2,2], i = 2,3,\ldots, n \\
\end{split}
\end{equation}

The second multiobjective task $T_{2}$ is defined as follows:
\begin{equation}\label{eq-t32}
\begin{split}
& \min f_{1}(\mathbf{x})= x_{1},\\
& \min f_{2}(\mathbf{x})= q(\mathbf{x})(1-\sqrt{\frac{x_{1}}{q(\mathbf{x})}}), \\
& q(\mathbf{x}) = 21 + e -20 exp(-0.2\sqrt{\frac{1}{n-1}\sum_{i=2}^{n}x_{i}^{2}}) \\
&   -exp(\frac{1}{n-1}\sum_{i=2}^{n}\cos(2 \pi x_{i}))\\
& x_{1} \in [0,1], x_{i} \in [-1,1], i = 2,3,\ldots, n \\
\end{split}
\end{equation}

For this test problem, the number of decision variables $n$ is set to 50
for both $T_{1}$ and $T_{2}$, and the similarity between $T_{1}$
and $T_{2}$ is 0.07.
The PS and PF of
$T_{1}$ are given as follows:
\begin{equation}\label{eq-pf31}
\begin{split}
& PS:  x_{1} \in [0,1],  x_{i}=0, i= 2,3,\ldots, 50 \\
& PF:  f_{1}^{2}+ f_{2}^{2}=1, f_{1} \geq 0, f_{2} \geq 0\\
\end{split}
\end{equation}
The PS and PF of
$T_{2}$ are given as follows:
\begin{equation}\label{eq-pf32}
\begin{split}
& PS: x_{1} \in [0,1],  x_{i}=0, i= 2,3,\ldots, 50 \\
& PF: f_{2}=1-\sqrt{f_{1}}, 0 \leq f_{1} \leq 1\\
\end{split}
\end{equation}
\\

4) \textbf{Partial Intersection with High Similarity (PIHS)}

The first multiobjective task $T_{1}$ is defined as follows:
\begin{equation}\label{eq-t41}
\begin{split}
& \min f_{1}(\mathbf{x})= x_{1},\\
& \min f_{2}(\mathbf{x})= q(\mathbf{x})(1-\sqrt{\frac{x_{1}}{q(\mathbf{x})}}), \\
& q(\mathbf{x}) =1+ \sum_{i=2}^{n}x_{i}^{2}, \\
& x_{1} \in [0,1], x_{i} \in [-100,100], i = 2,3,\ldots, n \\
\end{split}
\end{equation}

The second multiobjective task $T_{2}$ is defined as follows:
\begin{equation}\label{eq-t42}
\begin{split}
& \min f_{1}(\mathbf{x})= x_{1},\\
& \min f_{2}(\mathbf{x})= q(\mathbf{x})(1-\sqrt{\frac{x_{1}}{q(\mathbf{x})}}), \\
&  q(\mathbf{x}) = 1+\sum_{i=2}^{n} (z_{i}^{2} - 10 \cos(2 \pi z_{i}) + 10), \\
& (z_{2}, z_{3}, \ldots z_{n})^{\mathrm{T}} = (x_{2}, x_{3}, \ldots, x_{n})^{\mathrm{T}} - \textbf{s}_{ph2}, \\
& x_{1} \in [0,1], x_{i} \in [-100,100], i = 2,3,\ldots, n \\
\end{split}
\end{equation}

For this test problem, the number of decision variables $n$ is set to 50
for both $T_{1}$ and $T_{2}$, and the similarity between $T_{1}$
and $T_{2}$ is 0.99.
The PS and PF of
$T_{1}$ are given as follows:
\begin{equation}\label{eq-pf41}
\begin{split}
& PS:  x_{1} \in [0,1],  x_{i}=0, i= 2,3,\ldots, 50 \\
& PF: f_{2}=1-\sqrt{f_{1}}, 0 \leq f_{1} \leq 1\\
\end{split}
\end{equation}
The PS and PF of
$T_{2}$ are given as follows:
\begin{equation}\label{eq-pf42}
\begin{split}
& PS: x_{1} \in [0,1], (x_{2}, x_{3}, \ldots, x_{50})^{\mathrm{T}} = \textbf{s}_{ph2} \\
& PF: f_{2}=1-\sqrt{f_{1}}, 0 \leq f_{1} \leq 1\\
\end{split}
\end{equation}
\\

5) \textbf{Partial Intersection with Medium Similarity (PIMS)}

The first multiobjective task $T_{1}$ is defined as follows:
 \begin{equation}\label{eq-t51}
\begin{split}
& \min f_{1}(\mathbf{x}) = q(\mathbf{x})\cos(\frac{\pi x_{1}}{2}), \\
& \min f_{2}(\mathbf{x}) = q(\mathbf{x})\sin(\frac{\pi x_{1}}{2}), \\
& q(\mathbf{x}) = 1+\sum_{i=2}^{n} z_{i}^{2}  \\
& (z_{2}, z_{3}, \ldots z_{n})^{\mathrm{T}} = \mathbf{M}_{pm1}((x_{2}, x_{3}, \ldots, x_{n})^{\mathrm{T}} - \textbf{s}_{pm1}), \\
& x_{i} \in [0,1], i=1, 2,3,\ldots, n \\
\end{split}
\end{equation}

The second multiobjective task $T_{2}$ is defined as follows:
\begin{equation}\label{eq-t52}
\begin{split}
& \min f_{1}(\mathbf{x})= x_{1},\\
& \min f_{2}(\mathbf{x})= q(\mathbf{x})(1-(\frac{x_{1}}{q(\mathbf{x})})^{2}), \\
& q(\mathbf{x}) = 1+\sum_{i=2}^{n} (z_{i}^{2} - 10 \cos(2 \pi z_{i}) + 10), \\
& (z_{2}, z_{3}, \ldots z_{n})^{\mathrm{T}} = \mathbf{M}_{pm2}(x_{2}, x_{3}, \ldots, x_{n})^{\mathrm{T}}, \\
& x_{i} \in [0,1], i = 1, 2,3,\ldots, n \\
\end{split}
\end{equation}

For this test problem, the number of decision variables $n$ is set to 50
for both $T_{1}$ and $T_{2}$, and the similarity between $T_{1}$
and $T_{2}$ is 0.55.
The PS and PF of
$T_{1}$ are given as follows:
\begin{equation}\label{eq-pf51}
\begin{split}
& PS: x_{1} \in [0,1], (x_{2}, x_{3}, \ldots, x_{50})^{\mathrm{T}} = \textbf{s}_{pm1}\\
& PF: f_{1}^{2}+ f_{2}^{2}=1, f_{1} \geq 0, f_{2} \geq 0\\
\end{split}
\end{equation}
The PS and PF of
$T_{2}$ are given as follows:
\begin{equation}\label{eq-pf52}
\begin{split}
& PS: x_{1} \in [0,1], x_{i}=0, i= 2,3,\ldots, 50 \\
& PF: f_{2}=1-f_{1}^{2}, 0 \leq f_{1} \leq 1\\
\end{split}
\end{equation}
\\

6) \textbf{Partial Intersection with Low Similarity (PILS)}

The first multiobjective task $T_{1}$ is defined as follows:
 \begin{equation}\label{eq-t61}
\begin{split}
& \min f_{1}(\mathbf{x}) = q(\mathbf{x})\cos(\frac{\pi x_{1}}{2}), \\
& \min f_{2}(\mathbf{x}) = q(\mathbf{x})\sin(\frac{\pi x_{1}}{2}), \\
& q(\mathbf{x}) = 2 + \frac{1}{4000} \sum_{i=2}^{n}x_{i}^{2} - \prod_{i=2}^{n} \cos{(\frac{x_{i}}{\sqrt{i-1}})}\\
& x_{1} \in [0,1], x_{i} \in [-50,50], i= 2,3,\ldots, n \\
\end{split}
\end{equation}

The second multiobjective task $T_{2}$ is defined as follows:
 \begin{equation}\label{eq-t62}
\begin{split}
& \min f_{1}(\mathbf{x}) = q(\mathbf{x})\cos(\frac{\pi x_{1}}{2}), \\
& \min f_{2}(\mathbf{x}) = q(\mathbf{x})\sin(\frac{\pi x_{1}}{2}), \\
& q(\mathbf{x}) = 21 + e -20 exp(-0.2\sqrt{\frac{1}{n-1}\sum_{i=2}^{n}z_{i}^{2}})\\
&  -exp(\frac{1}{n-1}\sum_{i=2}^{n}\cos(2 \pi z_{i}))  \\
& (z_{2}, z_{3}, \ldots z_{n})^{\mathrm{T}} = (x_{2}, x_{3}, \ldots, x_{n})^{\mathrm{T}}-\mathbf{s}_{pl2}, \\
& x_{1} \in [0,1], x_{i} \in [-100,100], i= 2,3,\ldots, n \\
\end{split}
\end{equation}

For this test problem, the number of decision variables $n$ is set to 50
for both $T_{1}$ and $T_{2}$, and the similarity between $T_{1}$
and $T_{2}$ is 0.002.
The PS and PF of
$T_{1}$ are given as follows:
\begin{equation}\label{eq-pf61}
\begin{split}
& PS: x_{1} \in [0,1], x_{i}=0, i= 2,3,\ldots, 50 \\
& PF: f_{1}^{2}+ f_{2}^{2}=1, f_{1} \geq 0, f_{2} \geq 0\\
\end{split}
\end{equation}
The PS and PF of
$T_{2}$ are given as follows:
\begin{equation}\label{eq-pf62}
\begin{split}
& PS: x_{1} \in [0,1], (x_{2}, x_{3}, \ldots, x_{50})^{\mathrm{T}} = \textbf{s}_{pl2}\\
& PF: f_{1}^{2}+ f_{2}^{2}=1, f_{1} \geq 0, f_{2} \geq 0\\
\end{split}
\end{equation}
\\


7) \textbf{No Intersection with High Similarity (NIHS)}

The first multiobjective task $T_{1}$ is defined as follows:
\begin{equation}\label{eq-t71}
\begin{split}
& \min f_{1}(\mathbf{x}) = q(\mathbf{x})\cos(\frac{\pi x_{1}}{2}), \\
& \min f_{2}(\mathbf{x}) = q(\mathbf{x})\sin(\frac{\pi x_{1}}{2}), \\
&  q(\mathbf{x})= 1+\sum_{i=2}^{n-1} (100(x_{i}^{2}-x_{i+1})^{2} + (1-x_{i})^{2}),\\
& x_{i} \in [0,1], x_{i} \in [-80,80], i= 2,3,\ldots, n \\
\end{split}
\end{equation}

The second multiobjective task $T_{2}$ is defined as follows:
\begin{equation}\label{eq-t72}
\begin{split}
& \min f_{1}(\mathbf{x})= x_{1},\\
& \min f_{2}(\mathbf{x})= q(\mathbf{x})(1-\sqrt{\frac{x_{1}}{q(\mathbf{x})}}), \\
&   q(\mathbf{x}) = 1+\sum_{i=2}^{n}x_{i}^{2}, \\
& x_{1} \in [0,1], x_{i} \in [-80,80], i = 2,3,\ldots, n \\
\end{split}
\end{equation}

For this test problem, the number of decision variables $n$ is set to 50
for both $T_{1}$ and $T_{2}$, and the similarity between $T_{1}$
and $T_{2}$ is 0.94.
The PS and PF of
$T_{1}$ are given as follows:
\begin{equation}\label{eq-pf71}
\begin{split}
& PS:  x_{1} \in [0,1],  x_{i}=1, i= 2,3,\ldots, 50 \\
& PF:  f_{1}^{2}+ f_{2}^{2}=1, f_{1} \geq 0, f_{2} \geq 0\\
\end{split}
\end{equation}
The PS and PF of
$T_{2}$ are given as follows:
\begin{equation}\label{eq-pf72}
\begin{split}
& PS: x_{1} \in [0,1],  x_{i}=0, i= 2,3,\ldots, 50 \\
& PF: f_{2}=1-\sqrt{f_{1}}, 0 \leq f_{1} \leq 1\\
\end{split}
\end{equation}
\\

8) \textbf{No Intersection with Medium Similarity (NIMS)}

The first multiobjective task $T_{1}$ is defined as follows:
\begin{equation}\label{eq-t81}
\begin{split}
& \min f_{1}(\mathbf{x}) = q(\mathbf{x})\cos(\frac{\pi x_{1}}{2})\cos(\frac{\pi x_{2}}{2}), \\
& \min f_{2}(\mathbf{x}) = q(\mathbf{x})\cos(\frac{\pi x_{1}}{2})\sin(\frac{\pi x_{2}}{2}), \\
& \min f_{3}(\mathbf{x}) = q(\mathbf{x})\sin(\frac{\pi x_{1}}{2}), \\
& q(\mathbf{x})= 1+\sum_{i=3}^{n-1} (100(x_{i}^{2}-x_{i+1})^{2} + (1-x_{i})^{2}),\\
& x_{1} \in [0,1], x_{2} \in [0,1], x_{i} \in [-20,20], i = 3,4,\ldots, n \\
\end{split}
\end{equation}

The second multiobjective task $T_{2}$ is defined as follows:
\begin{equation}\label{eq-t82}
\begin{split}
& \min f_{1}(\mathbf{x})=\frac{1}{2} (x_{1} + x_{2}),\\
& \min f_{2}(\mathbf{x})= q(\mathbf{x})(1-(\frac{x_{1} + x_{2}}{2 \cdot q(\mathbf{x})})^{2}), \\
& q(\mathbf{x}) =1+  \sum_{i=3}^{n}z_{i}^{2}  \\
& (z_{3}, z_{4}, \ldots z_{n})^{\mathrm{T}} = \mathbf{M}_{nm2}(x_{3}, x_{4}, \ldots, x_{n})^{\mathrm{T}}, \\
& x_{1} \in [0,1], x_{2} \in [0,1], x_{i} \in [-20,20], i = 3,4,\ldots, n \\
\end{split}
\end{equation}

For this test problem, the number of decision variables $n$ is set to 20
for both $T_{1}$ and $T_{2}$, and the similarity between $T_{1}$
and $T_{2}$ is 0.51.
The PS and PF of
$T_{1}$ are given as follows:
\begin{equation}\label{eq-pf81}
\begin{split}
& PS: x_{1} \in [0,1], x_{2} \in [0,1], x_{i}=1, i=3,4,\ldots, 20 \\
& PF: f_{1}^{2}+ f_{2}^{2}+ f_{3}^{2}=1, f_{1} \geq 0, f_{2} \geq 0, f_{3} \geq 0\\
\end{split}
\end{equation}
The PS and PF of
$T_{2}$ are given as follows:
\begin{equation}\label{eq-pf82}
\begin{split}
& PS: x_{1} \in [0,1], x_{2} \in [0,1], x_{i}=0, i=3,4,\ldots, 20 \\
& PF: f_{2}=1-f_{1}^{2}, 0 \leq f_{1} \leq 1\\
\end{split}
\end{equation}
\\

9) \textbf{No Intersection with Low Similarity (NILS)}

The first multiobjective task $T_{1}$ is defined as follows:
\begin{equation}\label{eq-t91}
\begin{split}
& \min f_{1}(\mathbf{x}) = q(\mathbf{x})\cos(\frac{\pi x_{1}}{2})\cos(\frac{\pi x_{2}}{2}), \\
& \min f_{2}(\mathbf{x}) = q(\mathbf{x})\cos(\frac{\pi x_{1}}{2})\sin(\frac{\pi x_{2}}{2}), \\
& \min f_{3}(\mathbf{x}) = q(\mathbf{x})\sin(\frac{\pi x_{1}}{2}), \\
& q(\mathbf{x}) = 2 + \frac{1}{4000} \sum_{i=3}^{n}z_{i}^{2} - \prod_{i=3}^{n} \cos{(\frac{z_{i}}{\sqrt{i-2}})}\\
& (z_{3}, z_{4}, \ldots z_{n})^{\mathrm{T}} = (x_{3}, x_{4}, \ldots, x_{n})^{\mathrm{T}}-\mathbf{s}_{nl1}, \\
& x_{1} \in [0,1], x_{2} \in [0,1], x_{i} \in [-50,50], i = 3,4,\ldots, n \\
\end{split}
\end{equation}

The second multiobjective task $T_{2}$ is defined as follows:
\begin{equation}\label{eq-t92}
\begin{split}
& \min f_{1}(\mathbf{x})= \frac{1}{2} (x_{1} + x_{2}),\\
& \min f_{2}(\mathbf{x})= q(\mathbf{x})(1-(\frac{x_{1} + x_{2}}{2 \cdot q(\mathbf{x})})^{2}), \\
&  q(\mathbf{x}) = 21 + e -20 exp(-0.2\sqrt{\frac{1}{n-2}\sum_{i=3}^{n}x_{i}^{2}})\\
&  -exp(\frac{1}{n-2}\sum_{i=3}^{n}\cos(2 \pi x_{i}))\\
& x_{1} \in [0,1], x_{2} \in [0,1], x_{i} \in [-100,100], i = 3,4,\ldots, n \\
\end{split}
\end{equation}

For this test problem, the number of decision variables $n$ is set to 25
for $T_{1}$ and 50 for $T_{2}$, and the similarity between $T_{1}$
and $T_{2}$ is 0.001.
The PS and PF of
$T_{1}$ are given as follows:
\begin{equation}\label{eq-pf91}
\begin{split}
& PS: x_{1} \in [0,1], x_{2} \in [0,1], (x_{3},x_{4},\ldots,x_{25})^{\mathrm{T}}= \mathbf{s}_{nl1}\\
& PF: f_{1}^{2}+ f_{2}^{2}+ f_{3}^{2}=1, f_{1} \geq 0, f_{2} \geq 0, f_{3} \geq 0\\
\end{split}
\end{equation}
The PS and PF of
$T_{2}$ are given as follows:
\begin{equation}\label{eq-pf92}
\begin{split}
& PS: x_{1} \in [0,1], x_{2} \in [0,1], x_{i}=0, i=3,4,\ldots, 50 \\
& PF: f_{2}=1-f_{1}^{2}, 0 \leq f_{1} \leq 1\\
\end{split}
\end{equation}
\\

\begin{figure}[htbp]
\centering
    \includegraphics[scale=0.45]{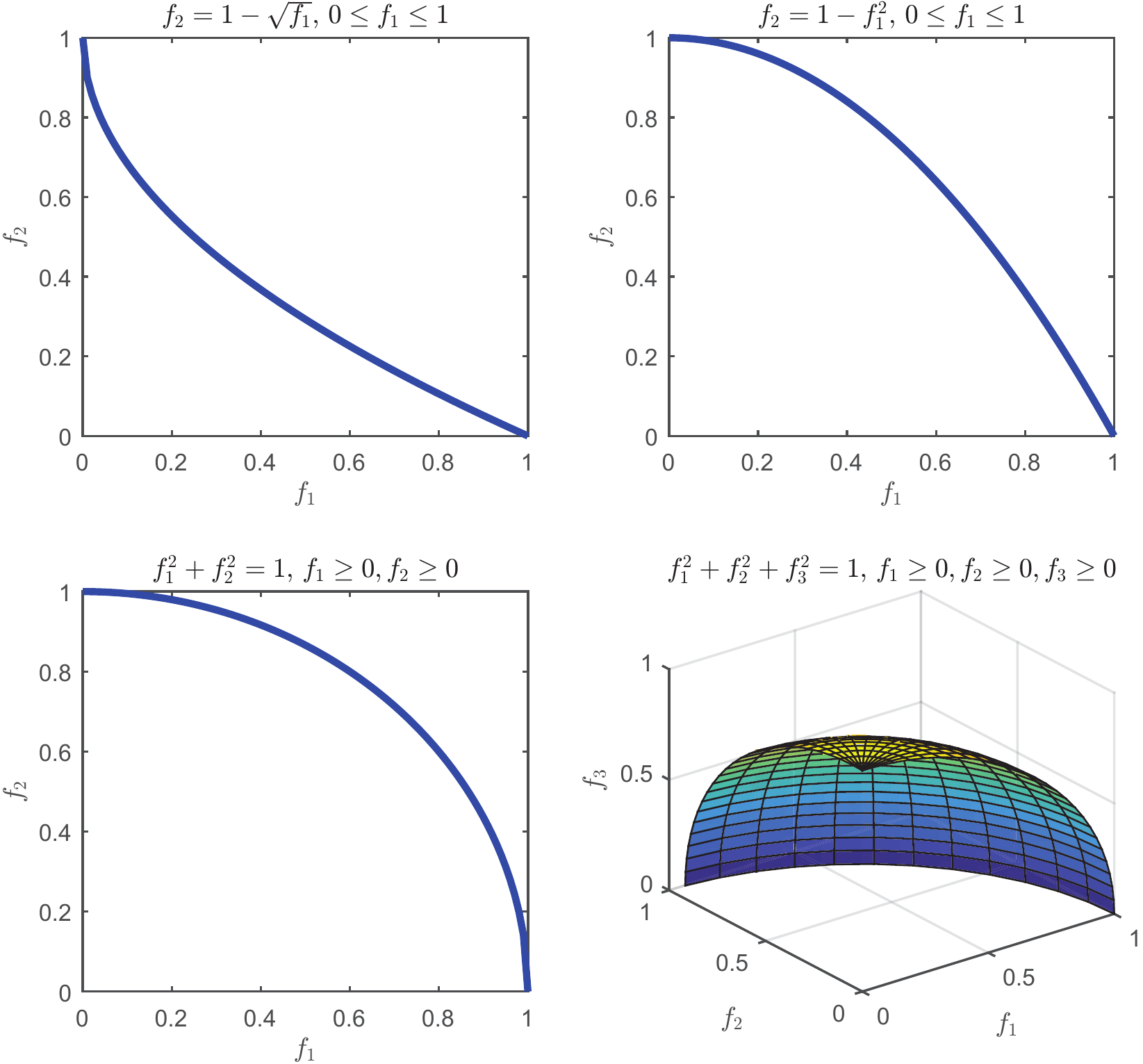}
\caption{Four different kinds of Pareto fronts involved in the proposed test problems.} \label{fig-pfs}
\end{figure}

Table \ref{tab-summary} summarizes the proposed test problems together with their properties, where $sim(T_{1},T_{2})$
denotes the similarity between two tasks in a test problem. Fig. \ref{fig-pfs}
shows four kinds of the Pareto fronts of the multiobjective optimization tasks in the test problems.

\begin{table*}[htbp]
\renewcommand{\arraystretch}{1.15}
\centering
\caption{Summary of the proposed test problems for Evolutionary Multiobjective Multitasking.}\label{tab-summary} \footnotesize \tabcolsep = 15pt
\begin{tabular}{cccccc}
\toprule
\multirow{2}{*}{Problem} &  $sim$ &   Task  &    \multirow{2}{*}{Pareto Set} &  \multirow{2}{*}{Pareto Front}  & \multirow{2}{*}{Properties}\\

 &  $(T_{1}, T_{2})$ &    No.               \\
\midrule

\multirow{4}{*}{CIHS}&\multirow{4}{*}{0.97} & \multirow{2}{*}{$T_{1}$}   &  $x_{1} \in [0,1]$,  &$f_{1}^{2}+f_{2}^{2}=1$, & concave, unimodal,\\
&&& $x_{i}=0, i=2:50$ &  $f_{1}\geq 0, f_{2}\geq 0$ & separable\\
\cmidrule{3-6}
&                                           & \multirow{2}{*}{$T_{2}$}  &   $x_{1} \in [0,1]$, & $f_{2}=1-f_{1}^{2}$, &concave, unimodal,\\
& && $x_{i}=0,i=2:50$ &  $0 \leq f_{1}\leq 1$& separable\\

\midrule

\multirow{4}{*}{CIMS}&\multirow{4}{*}{0.52} & \multirow{2}{*}{$T_{1}$}   &   $x_{1} \in [0,1]$, & $f_{2}=1-f_{1}^{2}$,
&concave, multimodal,\\
  &&& $x_{i}=1,i=2:10$ &  $0 \leq f_{1}\leq 1$& nonseparable\\
\cmidrule{3-6}
&                                           & \multirow{2}{*}{$T_{2}$}   & $x_{1} \in [0,1]$,  & $f_{1}^{2}+f_{2}^{2}=1$, &concave, unimodal, \\
   &&& $(x_{2},\ldots, x_{10})^{\mathrm{T}}=\mathbf{s}_{cm2}$   & $f_{1}\geq 0, f_{2}\geq 0$ & nonseparable\\

\midrule

\multirow{4}{*}{CILS}&\multirow{4}{*}{0.07} & \multirow{2}{*}{$T_{1}$} & $x_{1} \in [0,1]$, & $f_{1}^{2}+f_{2}^{2}=1$, &concave, multimodal,\\
  &&& $x_{i}=0,i=2:50$  &  $f_{1}\geq 0, f_{2}\geq 0$& separable\\
\cmidrule{3-6}
&                                           & \multirow{2}{*}{$T_{2}$} &  $x_{1} \in [0,1]$, & $f_{2}=1-\sqrt{f_{1}}$,  &convex, multimodal,\\
 &&& $x_{i}=0,i=2:50$  &  $0 \leq f_{1}\leq 1$& nonseparable\\
\midrule

\multirow{4}{*}{PIHS}&\multirow{4}{*}{0.99} & \multirow{2}{*}{$T_{1}$}  & $x_{1} \in [0,1]$, & $f_{2}=1-\sqrt{f_{1}}$, &convex, unimodal,\\
 &&& $x_{i}=0,i=2:50$   &  $0 \leq f_{1}\leq 1$& separable\\
\cmidrule{3-6}
&                                           & \multirow{2}{*}{$T_{2}$} & $x_{1} \in [0,1]$, & $f_{2}=1-\sqrt{f_{1}}$,&convex, multimodal,\\
 &&& $(x_{2},\ldots,x_{50})^{\mathrm{T}}=\mathbf{s}_{ph2}$   &  $0 \leq f_{1}\leq 1$& separable\\
\midrule

\multirow{4}{*}{PIMS}&\multirow{4}{*}{0.55} & \multirow{2}{*}{$T_{1}$}  &  $x_{1} \in [0,1]$, &  $f_{1}^{2}+f_{2}^{2}=1$,&concave, unimodal, \\
 &&& $(x_{2},\ldots,x_{50})^{\mathrm{T}}=\mathbf{s}_{pm1}$   &  $f_{1}\geq 0, f_{2}\geq 0$ & nonseparable\\
\cmidrule{3-6}
&                                           & \multirow{2}{*}{$T_{2}$} &  $x_{1} \in [0,1]$, & $f_{2}=1-f_{1}^{2}$,&concave, multimodal,\\
 &&& $x_{i}=0,i=2:50$   &  $0 \leq f_{1}\leq 1$& nonseparable\\
\midrule

\multirow{4}{*}{PILS}&\multirow{4}{*}{0.002} & \multirow{2}{*}{$T_{1}$} & $x_{1} \in [0,1]$, &  $f_{1}^{2}+f_{2}^{2}=1$, &concave, multimodal,\\
 &&& $x_{i}=0,i=2:50$   &  $f_{1}\geq 0, f_{2}\geq 0$ & nonseparable\\
\cmidrule{3-6}
&                                           & \multirow{2}{*}{$T_{2}$} &  $x_{1} \in [0,1]$, &  $f_{1}^{2}+f_{2}^{2}=1$, &concave, multimodal,\\
 &&& $(x_{2},\ldots,x_{50})^{\mathrm{T}}=\mathbf{s}_{pl2}$   &  $f_{1}\geq 0, f_{2}\geq 0$ & nonseparable\\
\midrule

\multirow{4}{*}{NIHS}&\multirow{4}{*}{0.94} & \multirow{2}{*}{$T_{1}$}  &  $x_{1} \in [0,1]$, &  $f_{1}^{2}+f_{2}^{2}=1$, &concave, multimodal,\\
 &&& $x_{i}=1,i=2:50$   &  $f_{1}\geq 0, f_{2}\geq 0$ & nonseparable\\
\cmidrule{3-6}
&                                           & \multirow{2}{*}{$T_{2}$} &  $x_{1} \in [0,1]$, & $f_{2}=1-\sqrt{f_{1}}$,&convex, unimodal,\\
 &&& $x_{i}=0,i=2:50$   &  $0 \leq f_{1}\leq 1$& separable\\
\midrule

\multirow{4}{*}{NIMS}&\multirow{4}{*}{0.51} & \multirow{2}{*}{$T_{1}$}  &  $x_{1} \in [0,1], x_{2} \in [0,1]$, &  $\sum_{i=1}^{3}f_{i}^{2}=1$, &concave, multimodal, \\
 &&& $x_{i}=1,i=3:20$   &  $f_{i}\geq 0,i=1,2,3$ & nonseparable\\
\cmidrule{3-6}
&                                           & \multirow{2}{*}{$T_{2}$} &  $x_{1} \in [0,1],x_{2} \in [0,1]$, & $f_{2}=1-f_{1}^{2}$, &concave, unimodal,\\
 &&& $x_{i}=0,i=3:20$   & $0 \leq f_{1}\leq 1$ & nonseparable\\
\midrule

\multirow{4}{*}{NILS}&\multirow{4}{*}{0.001} & \multirow{2}{*}{$T_{1}$}  &  $x_{1} \in [0,1],x_{2} \in [0,1]$, &  $\sum_{i=1}^{3}f_{i}^{2}=1$, &concave, multimodal,\\
 &&& $(x_{3},\ldots,x_{25})^{\mathrm{T}}=\mathbf{s}_{nl1}$   &  $f_{i}\geq 0, i=1,2,3$ & nonseparable \\
\cmidrule{3-6}
&                                           & \multirow{2}{*}{$T_{2}$} &  $x_{1} \in [0,1],x_{2} \in [0,1]$, & $f_{2}=1-f_{1}^{2}$, &concave, multimodal, \\
 &&& $x_{i}=0,i=3:50$   & $0 \leq f_{1}\leq 1$ & nonseparable\\
\bottomrule
\end{tabular}
\end{table*}

\section{Performance Evaluation}
\label{sec-Performance Evaluation}
This section describes the process that will be followed to evaluate the performance
of the algorithm on the proposed test problems. All the test problems should
be treated as black-box problems, i.e., the analytic forms of these
problems are not known for the algorithms.

\subsection{Performance Metric}
\label{sec-Performance Metric}

The inverted generational distance (IGD) \cite{van1998multiobjective}
is used to evaluate the performance of an
algorithm on each task of the considered test problem.
Let $A$ be a set of nondominated objective vectors that are obtained for a task $T_{i}$ by the algorithm, and
$P^{*}$ be a set of uniformly distributed objective vectors over the PF of $T_{i}$.
$A$ and $P^{*}$ are first normalized using the maximum and minimum objective values among $P^{*}$,
then the metric IGD of the approximate set $A$ is calculated as:
\begin{equation}\label{eq-IGD}
IGD(A, P^{*}) = \frac{1}{|P^{*}|}\sqrt{\sum_{\mathbf{x} \in P^{*}}{(\min_{\mathbf{y} \in A}d(\mathbf{x},\mathbf{y}))^{2}}}
\end{equation}
where $d(\mathbf{x},\mathbf{y})$ is the Euclidean distance between $\mathbf{x}$ and $\mathbf{y}$ in the normalized objective space.
If $|P^{*}|$ is large enough to represent the PF, the $\text{IGD}(A, P^{*})$ could measure both convergence and diversity
of $A$ in a sense. A small value of $\text{IGD}(A, P^{*})$ means that $A$ must be very close to $P^{*}$ and cannot miss any part of $P^{*}$.

The data file of $P^{*}$ for each task and the source code of computing IGD
can also be available from the zip package provided online.

\subsection{Ranking of the MO-MFO Algorithms}
\label{sec-Ranking of the MO-MFO Algorithms}

For a multiobjective multitasking problem, several IGD values are obtained by each run of the algorithm, and one value
is for a task. Based on these IGD values, we provide a comprehensive criterion to evaluate the overall performance of a MO-MFO algorithm, so as to rank the algorithms on each test problem.

Suppose there are $K$ tasks $T_{1},T_{2}, \ldots, T_{K}$ in a test problem, and the IGD value
obtained by an algorithm in a run for $T_{i}$ is denoted as $I_{i}$, where $i=1,2,\ldots K$.
Moreover, suppose the average and standard deviation of the IGD values for $T_{i}$
are $\mu_{i}$ and $\sigma_{i}$ respectively, where $i=1,2,\ldots, K$.
The mean standard score (MSS) of the obtained IGD values for the test problem is computed as follows:
\begin{equation}\label{eq-mss}
MSS = \frac{1}{K}\sum_{i=1}^{K}\frac{I_{i}-\mu_{i}}{\sigma_{i}}
\end{equation}
In practice, $\mu_{i}$ and $\sigma_{i}$ will be calculated according to the
IGD results obtained by all the algorithms on the task $T_{i}$ in all runs.
MSS is used as a comprehensive criterion, and a smaller MSS value indicates better
overall performance of a MO-MFO algorithm on a test problem.

\subsection{Experimental Settings}
\label{sec-Experimental Settings}

\subsubsection{The maximum number of nondominated objective vectors in the approximate set}

After running the algorithm on a test problem, the IGD metric should be computed for each task of the problem.
The maximum number of nondominated objective vectors produced by the algorithm for computing the IGD (i.e., $|A|$)
should be:
\begin{itemize}
\item  100 for the task with two objectives.
\item  120 for the task with three objectives.
\end{itemize}

\subsubsection{The maximum number of function evaluations}
It is set to 200,000 for all the test problems.
Note that ``a function evaluation'' here means a calculation of the objective functions
of a task $T_{i}$, and the function evaluations on
different tasks are not distinguished.

\subsubsection{The number of independent runs}
Each algorithm should be run 30 times independently on each multiobjective multitasking test problem.

\subsubsection{The parameter settings for the algorithm}
The test problem NIMS or NILS consists
of a bi-objective task and a three-objective task. Whereas each of the remaining test problems,
consists of two bi-objective tasks.
The algorithm should use the same parameters for NIMS and NILS, and use the same ones
for the other test problems.

\section{Baseline Results}
\label{sec-Baseline Results}

This section presents the baseline results
obtained by a MO-MFO algorithm, referred to as MO-MFEA \cite{gupta2016multiobjective},
on the proposed test problems.
MO-MFEA will be degenerated into NSGA-II when there is only one task in the input.
To show the benefits of MO-MFO, the results of MO-MFEA are
also compared with those of NSGA-II. It is worth note that
NSGA-II is not a MO-MFO algorithm and it solves the two tasks in a test problem separately,
whereas MO-MFEA solves them simultaneously in a single population.

NSGA-II uses a population size of 100 for solving a single task, while MO-MFEA uses a population size of 200 for solving a test problem with two tasks.
The maximal number of function evaluations
on a task is set to 100,000 for NSGA-II.
To ensure a fair comparison, MO-MFEA uses
200,000 as the maximal number of function evaluations
for a test problem since it solves two tasks together at a time.
NSGA-II and MO-MFEA use the same simulated binary crossover
operator and polynomial mutation to produce the candidate individuals.
The parameters for crossover and mutation are shown in Table \ref{tab-parameters},
where $D$ is the dimensionality of the unified code representation \cite{gupta2016multifactorial}.
\begin{table}[htbp]
\renewcommand{\arraystretch}{1.0}
\centering
\caption{Parameters for crossover and mutation.}
\label{tab-parameters} \footnotesize \tabcolsep =12pt
\begin{tabular}{lc}
\toprule
Parameter   &   Value     \\
\midrule
Crossover probability ($p_{c}$) & 0.9 \\
Mutation probability ($p_{m}$) & $1/D$   \\
Distribution index for crossover ($\eta_{c}$)& 20  \\
Distribution index for mutation ($\eta_{m}$)& 20  \\
\bottomrule
\end{tabular}
\end{table}

\begin{table}[htbp]
\renewcommand{\arraystretch}{1}
\centering
\caption{The average and standard deviation (shown in the bracket) of IGD values obtained by MO-MFEA and NSGA-II. The significantly better IGD value for each task is highlighted in bold.}\label{tab-mainresults} \footnotesize \tabcolsep = 6.5pt
\begin{tabular}{ccccc}
\toprule
\multirow{2}{*}{Problem} &  $sim$ &                  Task              &  \multicolumn{2}{c}{IGD}\\
\cmidrule{4-5}
                         &  $(T_{1}, T_{2})$    &    No.               &   NSGA-II & MO-MFEA  \\
\midrule
\multirow{4}{*}{CIHS}&\multirow{4}{*}{0.97}&\multirow{2}{*}{$T_{1}$}&2.0234E-3&\textbf{3.9912E-4}\\
&&& \scriptsize{(5.1879E-4)}&\scriptsize{(9.7671E-5)}\\
\cmidrule{3-5}
&&\multirow{2}{*}{$T_{2}$}&4.3621E-3&\textbf{2.6491E-3}\\
&&& \scriptsize{(8.5407E-4)}&\scriptsize{(5.6744E-4)}\\
\midrule
\multirow{4}{*}{CIMS}&\multirow{4}{*}{0.52}&\multirow{2}{*}{$T_{1}$}&1.0045E-1&\textbf{4.5705E-2}\\
&&& \scriptsize{(7.5847E-2)}&\scriptsize{(6.5713E-2)}\\
\cmidrule{3-5}
&&\multirow{2}{*}{$T_{2}$}&2.2897E-2&\textbf{8.7723E-3}\\
&&& \scriptsize{(1.9082E-2)}&\scriptsize{(1.2175E-2)}\\
\midrule
\multirow{4}{*}{CILS}&\multirow{4}{*}{0.07}&\multirow{2}{*}{$T_{1}$}&2.5503E-1&\textbf{2.7105E-4}\\
&&& \scriptsize{(1.0821E-1)}&\scriptsize{(2.6731E-5)}\\
\cmidrule{3-5}
&&\multirow{2}{*}{$T_{2}$}&1.9905E-4&\textbf{1.8986E-4}\\
&&& \scriptsize{(6.513E-6)}&\scriptsize{(6.628E-6)}\\
\midrule
\multirow{4}{*}{PIHS}&\multirow{4}{*}{0.99}&\multirow{2}{*}{$T_{1}$}&1.1145E-3&\textbf{1.1029E-3}\\
&&& \scriptsize{(3.8107E-4)}&\scriptsize{(1.1949E-3)}\\
\cmidrule{3-5}
&&\multirow{2}{*}{$T_{2}$}&5.654E-2&\textbf{3.0406E-2}\\
&&& \scriptsize{(3.0369E-2)}&\scriptsize{(1.6188E-2)}\\
\midrule
\multirow{4}{*}{PIMS}&\multirow{4}{*}{0.55}&\multirow{2}{*}{$T_{1}$}&4.4903E-3&\textbf{2.6206E-3}\\
&&& \scriptsize{(1.6131E-3)}&\scriptsize{(1.1806E-3)}\\
\cmidrule{3-5}
&&\multirow{2}{*}{$T_{2}$}&1.5577E1&\textbf{1.0892E1}\\
&&& \scriptsize{(3.7002E0)}&\scriptsize{(3.962E0)}\\
\midrule
\multirow{4}{*}{PILS}&\multirow{4}{*}{0.002}&\multirow{2}{*}{$T_{1}$}&\textbf{2.7647E-4}&3.2401E-4\\
&&& \scriptsize{(1.2629E-4)}&\scriptsize{(8.8287E-5)}\\
\cmidrule{3-5}
&&\multirow{2}{*}{$T_{2}$}&6.3458E-1&\textbf{1.099E-2}\\
&&& \scriptsize{(8.8258E-4)}&\scriptsize{(2.0652E-3)}\\
\midrule
\multirow{4}{*}{NIHS}&\multirow{4}{*}{0.94}&\multirow{2}{*}{$T_{1}$}&3.1286E1&\textbf{1.5523E0}\\
&&& \scriptsize{(6.0683E1)}&\scriptsize{(2.4312E-2)}\\
\cmidrule{3-5}
&&\multirow{2}{*}{$T_{2}$}&7.8558E-4&\textbf{5.0224E-4}\\
&&& \scriptsize{(1.6655E-4)}&\scriptsize{(1.4331E-4)}\\
\midrule
\multirow{4}{*}{NIMS}&\multirow{4}{*}{0.51}&\multirow{2}{*}{$T_{1}$}&4.7021E-1&\textbf{2.792E-1}\\
&&& \scriptsize{(3.3048E-1)}&\scriptsize{(2.6437E-1)}\\
\cmidrule{3-5}
&&\multirow{2}{*}{$T_{2}$}&9.1944E-2&\textbf{2.8576E-2}\\
&&& \scriptsize{(8.8784E-2)}&\scriptsize{(4.7721E-2)}\\
\midrule
\multirow{4}{*}{NILS}&\multirow{4}{*}{0.001}&\multirow{2}{*}{$T_{1}$}&8.3588E-4&8.3483E-4\\
&&& \scriptsize{(5.4979E-5)}&\scriptsize{(6.2227E-5)}\\
\cmidrule{3-5}
&&\multirow{2}{*}{$T_{2}$}&\textbf{6.4226E-1}&6.4316E-1\\
&&& \scriptsize{(2.5715E-4)}&\scriptsize{(3.2324E-4)}\\
\bottomrule
\end{tabular}
\end{table}

\subsection{Results of IGD and MSS}
\label{Results of IGD and MSS}

Table \ref{tab-mainresults} shows the average IGD obtained by MO-MFEA and NSGA-II on nine test problems, where
the standard deviation values are also reported.
To test the difference for statistical
significance, the Wilcoxon signed-rank
test at a 5\% significance level is conducted on the
IGD values obtained by MO-MFEA and NSGA-II for each multiobjective task, and the significantly
better IGD value for each task is shown in bold.

From Table \ref{tab-mainresults}, except for PILS and NILS, MO-MFEA performs significantly better
than NSGA-II on both the two tasks in a test problem. For some cases, e.g., $T_{1}$ of CILS and $T_{1}$ of NIHS,
MO-MFEA even outperforms NSGA-II by a large margin.
For PILS, although MO-MFEA is a little worse than NSGA-II on $T_{1}$, it performs much better than NSGA-II
on $T_{2}$. As for NILS, MO-MFEA and NSGA-II achieve very close performance on both $T_{1}$
and $T_{2}$.

\begin{figure*}[htbp]
\centering
    \includegraphics[scale=0.62]{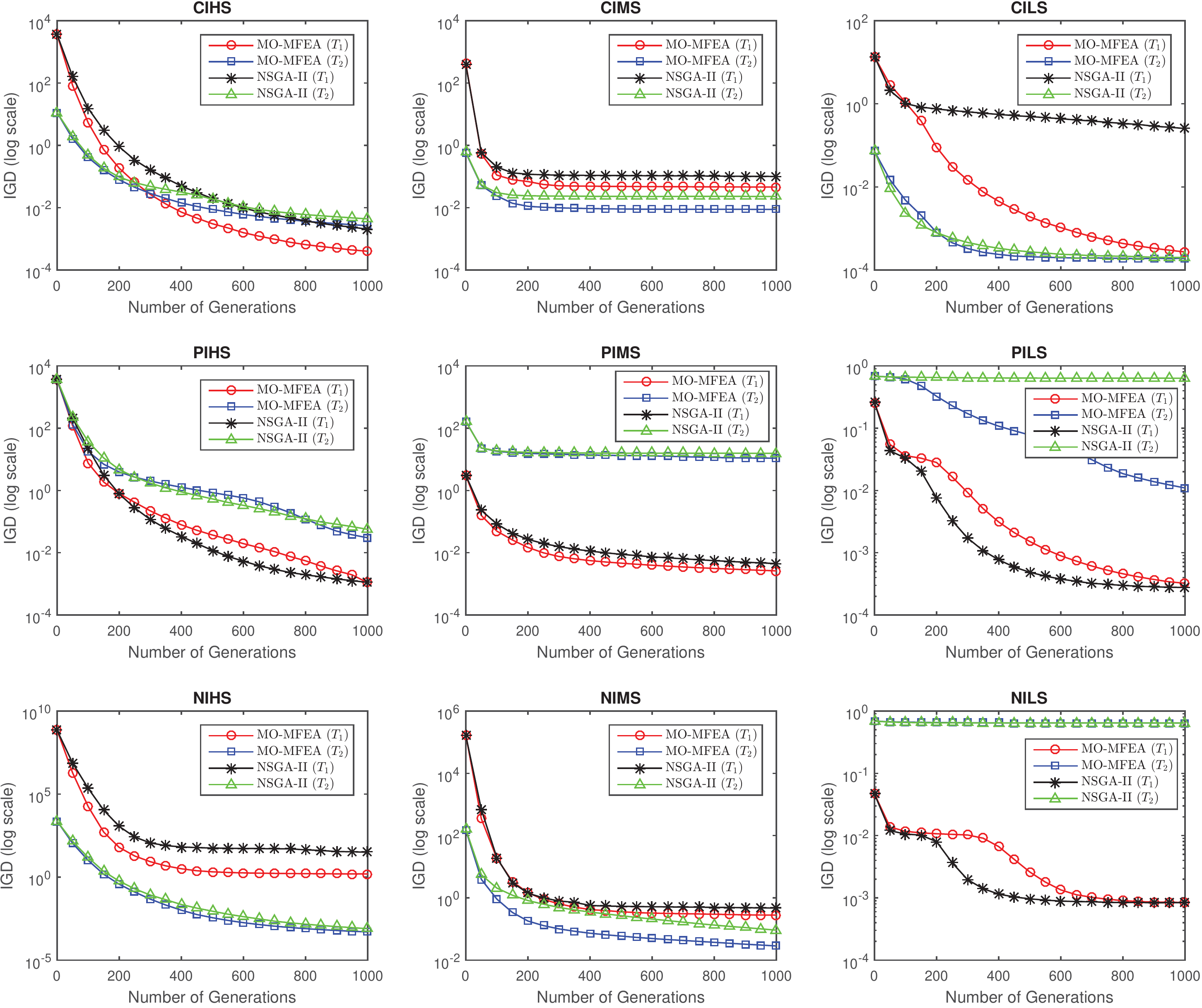}
\caption{Convergence curves of the average IGD (over 30 runs) obtained with the number of generations for MO-MFEA
and NSGA-II on the proposed test problems.} \label{fig-nsgaiiconv}
\end{figure*}

\begin{table}[htbp]
\renewcommand{\arraystretch}{1.2}
\centering
\caption{The difference between the average MSS obtained by MO-MFEA and NSGA-II.}\label{tab-mssresults} \footnotesize \tabcolsep = 4pt
\begin{tabular}{ccc}
\toprule
Problem &    $sim(T_{1}, T_{2})$   &   MSS Difference         \\

\midrule

COHS& 0.97& -1.672\\
COMS& 0.52& -0.764\\
COLS& 0.07& -1.43\\
POHS& 0.99& -0.48\\
POMS& 0.55& -1.073\\
POLS& 0.002& -0.787\\
NOHS& 0.94& -1.001\\
NOMS& 0.51& -0.71\\
NOLS& 0.001& 0.832\\
\bottomrule
\end{tabular}
\end{table}

Table \ref{tab-mssresults} further reports
the difference between the average MSS obtained by
MO-MFEA and NSGA-II on each test problem.
Here $\mu_{i}$ and $\sigma_{i}$ are calculated according to the
IGD results of MO-MFEA and NSGA-II for $T_{i}$ in all runs.
As can be seen from Table \ref{tab-mssresults}, all the differences are smaller
than 0 except on NILS, which indicates that MO-MFEA achieves better overall performance than NSGA-II
on all the proposed test problems except NILS.

\subsection{Convergence Curves}
\label{sec-Convergence Curves}
Fig. \ref{fig-nsgaiiconv} shows
the evolution of average IGD with number of generations for MO-MFEA and NSGA-II
on each of the test problem.
From Fig. \ref{fig-nsgaiiconv},
MO-MFEA converges to the two Pareto fronts (for $T_{1}$ and $T_{2}$ respectively) faster than NSGA-II on most of test problems by solving the two multiobjective tasks in a test problem simultaneously.
For certain cases, NSGA-II shows a little faster convergence speed in the
early stage of search but is surpassed by MO-MFEA quickly, which is
particularly clear on CILS.

\subsection{Distribution of Solutions in the Objective Space}

\begin{figure*}[htbp]
\centering
    \includegraphics[scale=0.68]{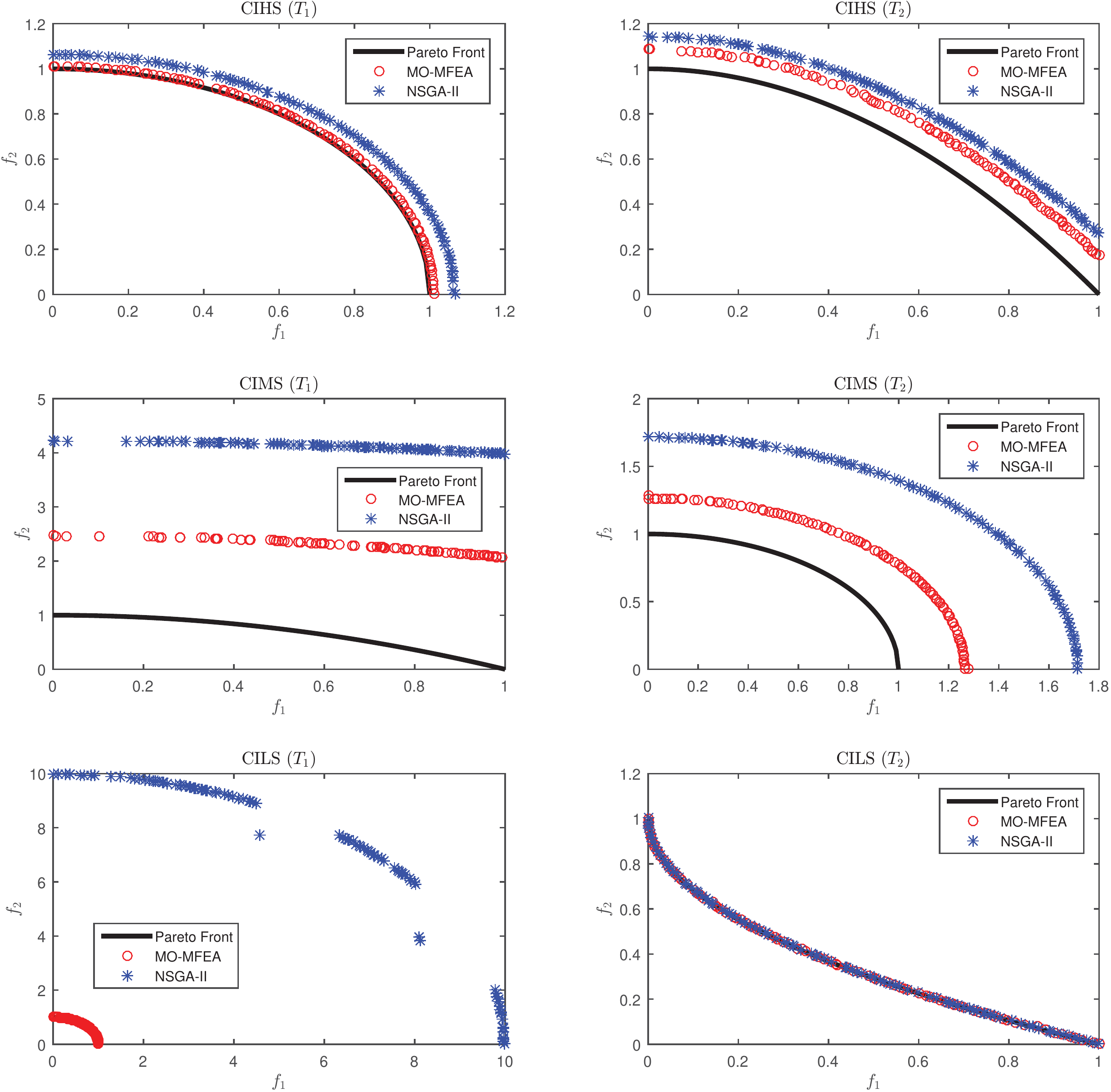}
\caption{Distribution of solutions obtained by MO-MFEA and NSGA-II in the objective space for each task.} \label{fig-d1}
\end{figure*}

\begin{figure*}[htbp]
\centering
    \includegraphics[scale=0.68]{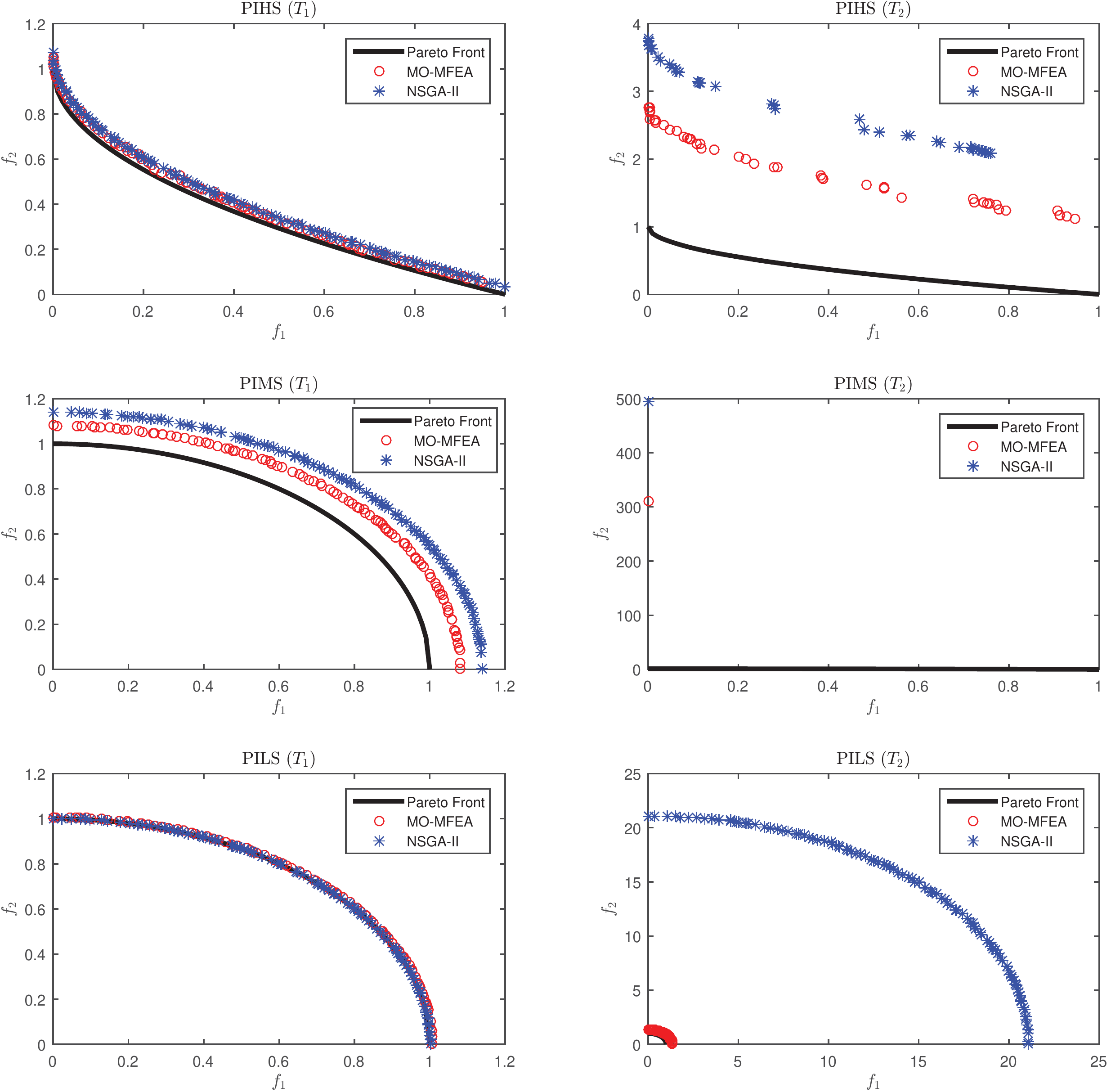}
\caption{Distribution of solutions obtained by MO-MFEA and NSGA-II in the objective space for each task (continued 1).} \label{fig-d2}
\end{figure*}

\begin{figure*}[htbp]
\centering
    \includegraphics[scale=0.68]{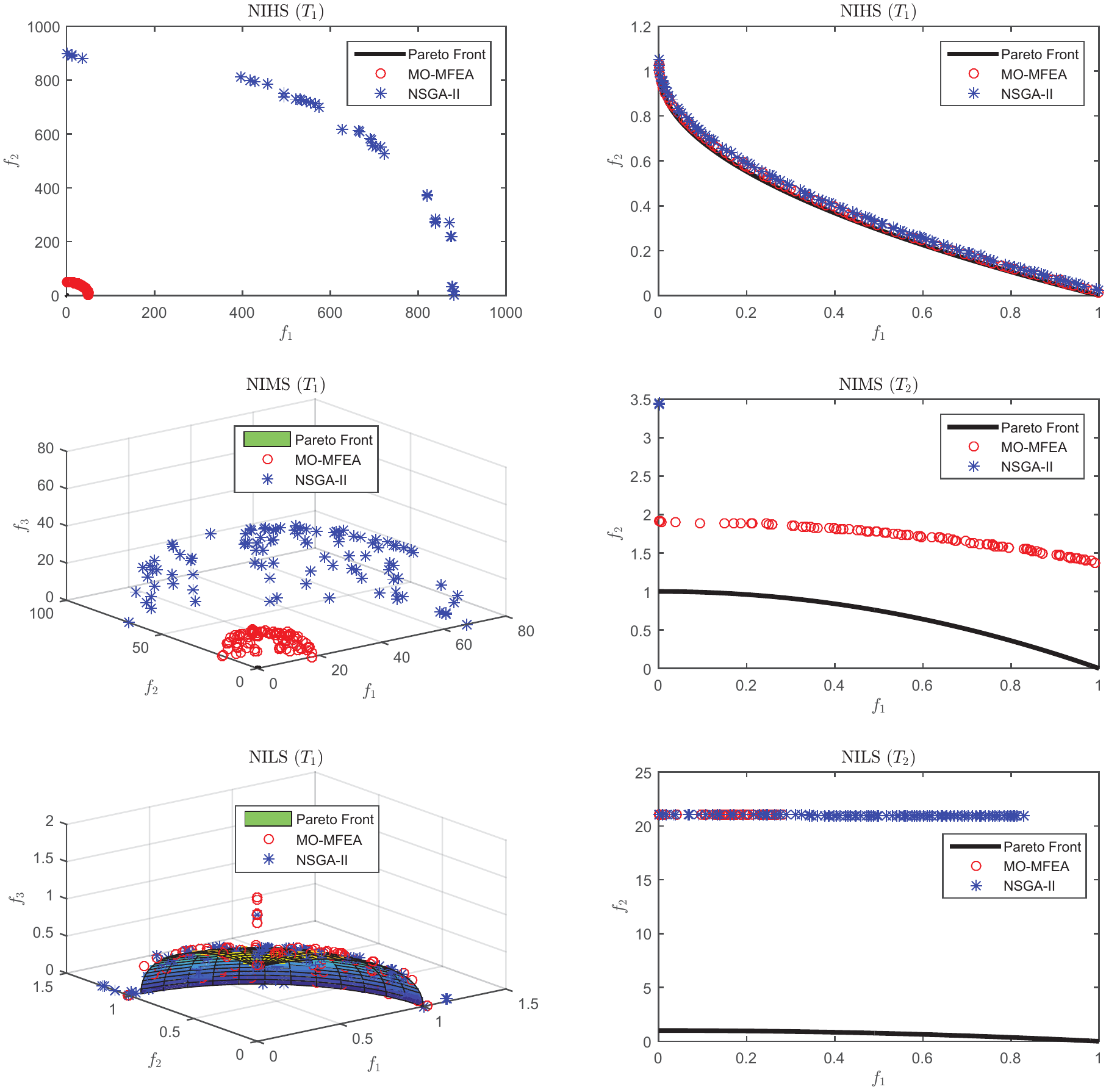}
\caption{Distribution of solutions obtained by MO-MFEA and NSGA-II in the objective space for each task (continued 2).} \label{fig-d3}
\end{figure*}

The distribution of nondominated objective vectors obtained by MO-MFEA and NSGA-II in a single run for each task is shown in Figs. \ref{fig-d1}, \ref{fig-d2}, and \ref{fig-d3},
in order to visually see how well the obtained approximate set is in the objective space.
This particular run is associated with
the result closest to the average IGD value.
It can be seen from these figures that MO-MFEA
can achieve a better PF approximation than NSGA-II almost on all
the concerned multiobjective optimization tasks.

\IEEEpeerreviewmaketitle



\end{document}